\documentclass[journal]{IEEEtran}
\usepackage{microtype}                 % use micro-typography (slightly more compact, better to read)
\PassOptionsToPackage{warn}{textcomp}  % to address font issues with \textrightarrow
\usepackage{tikz}
\usepackage{comment}
\usepackage{amsmath,amssymb} % define this before the line numbering.
\usepackage{color}
\usepackage[switch]{lineno}
\usepackage{comment}
\usepackage{bbm}
\usepackage{amsmath, xparse}
\usepackage{here}
\usepackage{xcolor}
\usepackage{subfig}
\usepackage{bm}
\usepackage{cite}
\usepackage{multirow}
\usepackage{ctable} 
\usepackage{arydshln}
\usepackage{algorithm}
\usepackage{algorithmic}
\usepackage{enumitem}
\newcommand\norm[1]{\left\lVert#1\right\rVert}
\newcommand{\specialcell}[2][c]{%
  \begin{tabular}[#1]{@{}c@{}}#2\end{tabular}}

\ifCLASSINFOpdf
\else
\fi
\begin{document}

%\title{Towards Diverse yet Controllable \\ Audio-Driven Facial Animation}
%\title{Exploring Motion Prior Space for Diverse yet \\ Controllable Audio-Driven Facial Animation}
%\title{FPA: Facial Parts Assembler for Diverse yet \\ Controllable Audio-Driven Facial Animation}
%\title{Sequential Latent Code Querying for Diverse Speech-Driven Facial Animation Synthesis}
%\title{Sequential Code Query Learning for Diverse Speech-Driven Facial Animation Synthesis}
\title{Diverse Code Query Learning for \\ Speech-Driven Facial Animation}

%\title{Sequential and Diverse Code Query Modeling for Speech-Driven Facial Animation Synthesis}

%Exploring Realistic Leg Dynamics for Orientation-Aware Action Transition Learning

% \author{Chunzhi~Gu,~\IEEEmembership{Member,~IEEE,}
%         Shigeru~Kuriyama,~\IEEEmembership{Member,~IEEE,}% <-this % stops a space
%         ~and~Katsuya~Hotta,~\IEEEmembership{Member,~IEEE}% <-this % stops a space
% \thanks{C. Gu* and S. Kuriyama are with the Department of Computer Science and Engineering, Toyohashi University of Technology, Toyohashi, Japan (e-mails: gu@cs.tut.ac.jp, sk@tut.jp).}% <-this % stops a space
% \thanks{K. Hotta is with the Faculty of Science and Engineering, Iwate University, Morioka, Japan (hotta@iwate-u.ac.jp).}
% \thanks{Manuscript received April xx, xx; revised August xx, xx, (Corresponding author: C. Gu.)}}

\author{Chunzhi~Gu,
        Shigeru~Kuriyama,% <-this % stops a space
        ~and~Katsuya~Hotta% <-this % stops a space
\thanks{C. Gu* and S. Kuriyama are with the Department of Computer Science and Engineering, Toyohashi University of Technology, Toyohashi, Japan (e-mails: gu@cs.tut.ac.jp, sk@tut.jp).}% <-this % stops a space
\thanks{K. Hotta is with the Faculty of Science and Engineering, Iwate University, Morioka, Japan (hotta@iwate-u.ac.jp).}
\thanks{Corresponding author: C. Gu.}}


\markboth{Journal of \LaTeX\ Class Files,~Vol.~14, No.~8, August~2015}%
{Shell \MakeLowercase{\textit{et al.}}: Bare Demo of IEEEtran.cls for IEEE Journals}



\maketitle

\begin{abstract}
Speech-driven facial animation aims to synthesize lip-synchronized 3D talking faces following the given speech signal. Prior methods to this task mostly focus on pursuing realism with deterministic systems, yet characterizing the potentially stochastic nature of facial motions has been to date rarely studied. While generative modeling approaches can easily handle the one-to-many mapping by repeatedly drawing samples, ensuring a diverse mode coverage of plausible facial motions on small-scale datasets remains challenging and less explored. In this paper, we propose predicting multiple samples conditioned on the same audio signal and then explicitly encouraging sample diversity to address diverse facial animation synthesis. Our core insight is to guide our model to explore the expressive facial latent space with a diversity-promoting loss such that the desired latent codes for diversification can be ideally identified. To this end, building upon the rich facial prior learned with vector-quantized variational auto-encoding mechanism, our model temporally queries multiple stochastic codes which can be flexibly decoded into a diverse yet plausible set of speech-faithful facial motions.  To further allow for control over different facial parts during generation, the proposed model is designed to predict different facial portions of interest in a sequential manner, and compose them to eventually form full-face motions. Our paradigm realizes both diverse and controllable facial animation synthesis in a unified formulation. We experimentally demonstrate that our method yields state-of-the-art performance both quantitatively and qualitatively, especially regarding sample diversity.

%In addition, limited audio-visual datasets impose more challenges.

%Since existing datasets do not one-to-many annotated 

%We formulate a VAE-based in-betweening learning framework 

\end{abstract}


\IEEEpeerreviewmaketitle

\section{Introduction}

\begin{figure*}[th]
\begin{center}
\includegraphics[width=1.0\linewidth]{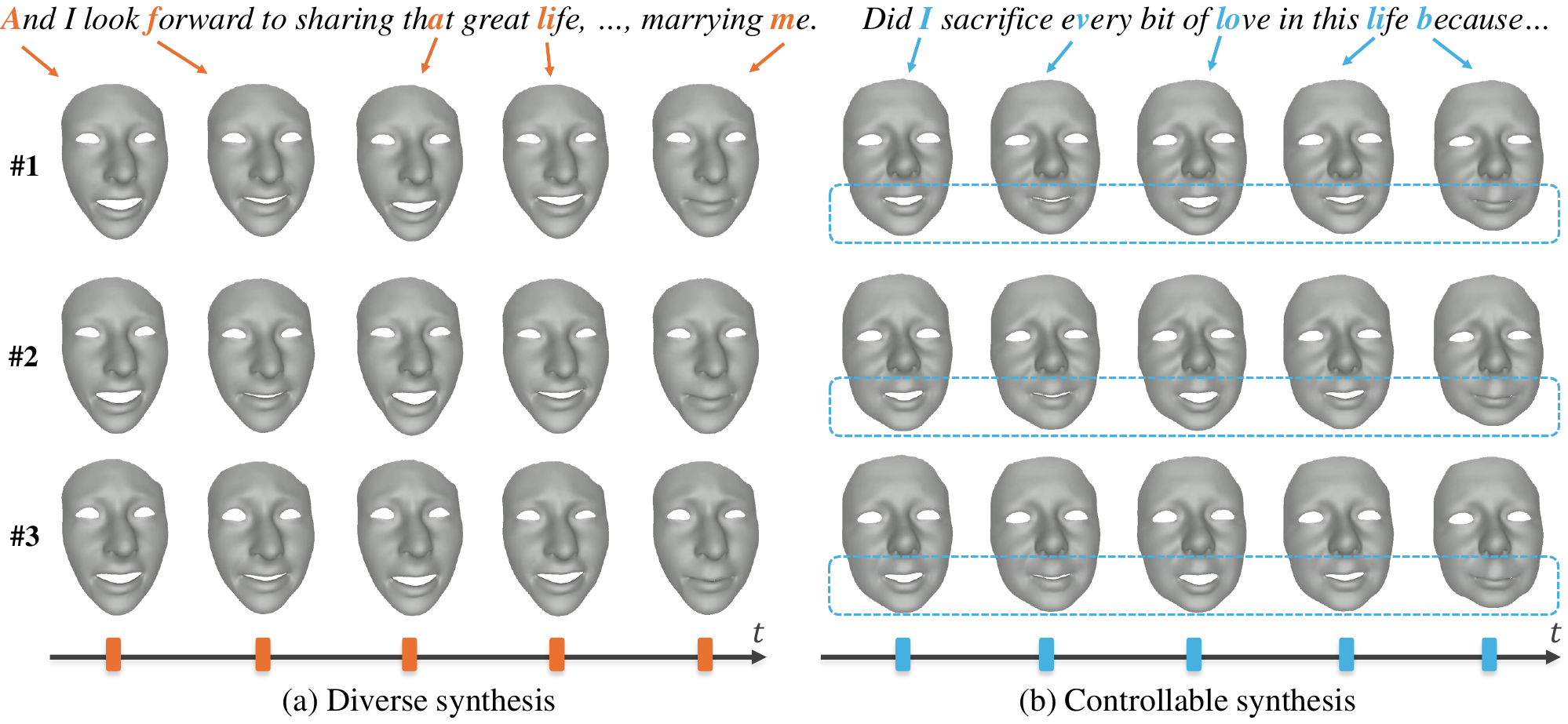}
\end{center}
\caption{\textbf{Diverse (a) and controllable (b) facial motion synthesis.} In the controllable setting (b), all samples have strictly fixed lip motions (blue dotted area) but with diverse upper-face variations. }
\label{fig:teaser}
\end{figure*}

\IEEEPARstart{S}{ynthesizing} 3D facial animations driven by speech audio has wide applications in gaming, filming, and virtual/augmented reality (VR/AR) \cite{zollhofer2018state} industries. The goal of this task is to capture the inner relationship between speech and facial movements to animate lip-synchronized 3D facial movements. In stark contrast to earlier efforts \cite{xu2013practical,edwards2016jali,taylor2012dynamic, massaro2012animated} that involve laborious manual tuning by technical animators, recent techniques focus on leveraging deep neural networks to learn facial dynamics conditioned on speech. 

Most current approaches \cite{fan2022faceformer,xing2023codetalker,richard2021meshtalk} follow deterministic generation, i.e., synthesizing only the most likely facial sequence, with carefully designed powerful learning schemes (e.g., periodic time encoding \cite{fan2022faceformer}). However, due to the potentially ill-posed nature of human behavior regarding personal styles or habits, the resulting talking facial movements should be multi-modal even given the same speech.  In principle, to characterize the complex one-to-many correlation, a naive strategy would be adopting conditional generative modeling for audio-conditioned facial motions. Multiple facial motion samples can then be derived by repeatedly sampling from the learned latent space. Considering the strong modeling capabilities, this direction has been mainly studied with diffusion-based approaches \cite{stan2023facediffuser,thambiraja20233diface}. However, such generation tends to induce highly similar samples. The reason can be attributed to the fact that the inference stage employs likelihood-based sampling, which causes the results to concentrate on the major data mode and can rarely access the minor modes in the solution space. This issue is even magnified by the scarcity of existing datasets where the paired audio-visual training samples are largely limited in amount and variation. More recently, Yang et al. \cite{yang2024probabilistic} introduced a probabilistic facial motion synthesis approach to addressing sample diversity, which is realized by performing different code sampling schemes (e.g., K-Nearest Neighbor) or even code manipulating (i.e., averaging) in the latent space. However, this strategy lacks clear and straightforward guidance to truly encourage diversity. %Another way to increase the inference diversity is by conditioning a style token that indicates the speaker's identity during training, which allows for the generation of multiple talking faces by switching identities. 

In this paper, we address the task of diverse speech-driven facial animation synthesis by proposing to explicitly generate multiple facial samples conditioned on the same input speech signal, and promote sample diversity. To this end, motivated by previous works \cite{xing2023codetalker,yang2024probabilistic}, we construct facial prior with vector-quantized variational auto-encoder (VQ-VAE), and predict the latent code as a proxy of facial motion itself to exploit the rich expressiveness of the discrete latent space for realism. As expecting the dataset to provide the required one-to-many paired ground-truth supervision with high diversity is infeasible, 
we therefore drive our model to explore ideal codes that constitute diverse samples with a diversity-promoting objective. \textit{In essence, our method can be understood as a diverse code querying mechanism using the speech signal, which forces the model to entirely discover different data modes for diversification.} Importantly, benefiting from the rich and valid geometry cues within the discrete facial prior, our model yields diverse yet plausible facial dynamics with high audio-fidelity. 

The second main challenge in stochastic generation is to allow for controllability over facial parts. As an application for digital avatars, one may desire multiple talking faces to share similar lip movements with diverse upper-face variations. To achieve this, we design our model to sequentially predict multiple samples for each facial portion, and eventually produce the full-face movements by composing these parts. The resulting partial facial priors are thus individually prepared to facilitate control. In addition to the stochastic synthesis capacity, our method further yields partially diversified results for the uncontrolled portion. Our modeling framework therefore realizes \textit{\textbf{C}ontrollable} (Fig. \ref{fig:teaser}(b)) and \textit{\textbf{D}iverse} (Fig. \ref{fig:teaser}(a)) talking \textit{\textbf{Face}} synthesis in a unified formulation, which we dub as \textbf{CDFace} in short. We perform extensive experiments to demonstrate that our model outperforms the state-of-the-art methods in synthesizing audio-faithful 3D facial motions while achieving controllability and high sample diversity.

Our contribution can be itemized as follows:
\begin{itemize}
\item  We propose a diverse code querying mechanism to identify target latent codes from vector-quantized prior space that yield diversified speech-conditioned 3D facial motion samples. 
\item  We design a unified model with sequential architecture to allow for controllable synthesis over facial parts. 
\item  We experimentally demonstrate state-of-the-art performance for facial animation synthesis against prior approaches, both quantitatively and qualitatively. 
\end{itemize}

%One of the key challenges is the lack of one-to-many paired ground-truth supervision with high diversity. 

\section{Related Work}
In this section, we first review previous speech-driven facial animation synthesis techniques. We then discuss some literature where quantized latent prior is involved. Finally, we discuss stochastic generation techniques for sequential modeling. 

\noindent	\textbf{Speech-Driven Facial Animation Synthesis.} 
As a branch of the long-lasting talking facial animation task  \cite{parke2008computer} in computer graphics, speech-driven facial animation is developed to condition the synthesis with audio to encourage speech synchronization. Early efforts in this field mostly follow the procedural modeling \cite{
xu2013practical,edwards2016jali,taylor2012dynamic, massaro2012animated} to exploit linguistic cues to form the lip motions. This involves a series of dedicated rules to understand the dependency between phonemes and visemes, such as the dominance
function in \cite{massaro2012animated} to predict facial control parameters. Despite the advantages of controllability and easy integration to other animating pipelines, the resulting complexity built for co-articulation can be laborious to tune. 

In contrast to the above methods, another line for this field developed learning-based strategies \cite{pei2006transferring,cao2005expressive,karras2017audio,taylor2017deep,zhou2018visemenet,liu2021geometry} to explore the mapping from speech to animation in a data-driven fashion. While early methods established conventional machine learning baselines, such as utilizing the graph structure \cite{cao2005expressive}, later works typically resort to deep learning to more effectively learn the audio-visual correlation. Talor et al. \cite{taylor2017deep} 
devised a continuous deep sliding window predictor to map the phonetic representation to visual speech. Zhou et al. \cite{zhou2018visemenet} proposed the VisemeNet that includes a three-stage Long-Short Term Memory (LSTM) network to model viseme animation curves by jointly considering facial landmarks, phoneme groups, and audio. More recent methods \cite{thambiraja2023imitator,xing2023codetalker,fan2022faceformer,sung2024laughtalk,li2024pose,chai2022personalized} adopt the Transformer-based network backbones to exploit the strong temporal learning capacity. For example, Fan et al. \cite{fan2022faceformer} devised periodic positional embedding to boost the generality to longer audio signals. Li et al. \cite{li2024pose} introduced a geometry-guided audio-vertices attention mechanism to reflect natural head poses during talking. 

These methods, however, do not model the stochastic nature of facial movements. In this regard, the most closely related methods to ours are \cite{stan2023facediffuser, yang2024probabilistic,sun2024diffpose} which enable the synthesis of multiple facial motions conditioned on the same audio. \cite{stan2023facediffuser} is a diffusion-based generative facial motion modeling approach, yet the samples generated during inference can mostly focus on the major data mode with low diversity. The framework in \cite{yang2024probabilistic} is a coarse-to-fine code manipulation strategy for stochastic talking face. However, it only works on large-scale datasets and thus cannot be easily applied to other common facial benchmarks where the data number is limited. We overcome these issues by designing a diverse code querying mechanism that enables us to explore rare data modes even from limited training data.

\noindent	\textbf{Quantized Latent Prior.} Due to the remarkable detail-preserving capability for visual media, vector-quantized (VQ) learning has received active attention for the past decade, particularly in the field of image generation \cite{huang2022unicolor,peng2021generating,zhang2024codebook,zhou2022towards}. By first constructing the quantized prior space that compactly stores rich texture features, the generation can then be cast as an auto-regressive code distribution modeling task using an additional prediction network. The basic VQ frameworks includes VQ-VAE \cite{yuvector2022,esser2021taming} and VQ-GAN \cite{razavi2019generating,van2017neural}. Each of these techniques utilizes a codebook whose tokens serve as the prior for the target data.

Besides the images, VQ priors can be naturally adapted to synthesize sequential data, such as human \cite{zhang2023generating}, facial \cite{xing2023codetalker,ng2022learning,yang2024probabilistic}, or holistic \cite{liu2024emage,yi2023generating} motion synthesis. Analogous to the case of images, the generation can be achieved by forecasting the discrete motion primitives within the codebook. Our method falls into this category and draws inspiration from these works in exploring the 
quantized latent prior, but differs centrally in the aim  to pursue diversity in addition to plausibility for facial motions with high audio fidelity.

\noindent	\textbf{Diverse Inference.} Exploring sample diversity in sequential generation has been primarily studied for human motion \cite{kim2024motion,yuan2020dlow,mao2021generating,xu2022diverse}, mostly devising post-hoc sampling strategies from generative models, but has been rarely surveyed for facial animation. In contrast to human motion generation, which generally entails texts or action conditions for global semantic guidance, the synthesis of facial movements given speech conditions demands precise frame-wise lip synchronization. As such, a direct adaptation of these models to facial motions can result in highly unfaithful results. In learning probabilistic talking faces, \cite{yang2024probabilistic} shares the closest motivation to ours. Specifically, it achieves diversity in facial movements by performing KNN/rejection-based sampling in the quantized latent space. Nevertheless, such simple sampling techniques lack a straightforward navigation to diversify the generation. Differently, our method is designed to explicitly drive the generation to a diverse configuration, with flexible controllability over facial parts. 
%Different from the 

%Regarding this, the most closely related work to ours is \cite{yang2024probabilistic}, 

%Furthermore, it has progressed to handle sequential data by regressing the per-frame token in the codebook. 

%Our methods differ mostly from these approaches in 

%such as  Anime Graph structure and a resulting node-searching method to model affective visual behavior, later 

%\noindent	\textbf{Diverse generation for sequential modeling.} 

\section{Method}
Given an input speech audio signal $\mathbf{A}$, the task of speech-driven facial animation synthesis aims to generate lip-synchronized facial motions $\mathbf{X}=[\mathbf{x}_1,\cdots,\mathbf{x}_T]$ with all $T$ timesteps, where $\mathbf{x}_t \in \mathbb{R}^{3V}$ refers to a 3D facial mesh with $V$ vertices. Precisely, $\mathbf{x}_t$ models the offset over a given neutral facial template $\mathbf{f} \in \mathbb{R}^{3V}$  as expression reference. The task can be thus converted to predict such displacements to shape the resulting talking face: $\mathbf{F}=[\mathbf{x}_1+\mathbf{f},\cdots,\mathbf{x}_T+\mathbf{f}]$. 

\subsection{Vector-Quantized Facial Prior Pair}
\label{sec:vq-vaes}
Instead of directly predicting facial expression itself, we draw inspiration from recent 3D face modeling schemes \cite{xing2023codetalker,tan2024flowvqtalker} by constructing a discrete low-dimensional facial prior to store high-quality visual textures for facial geometry with VQ-VAE.

\noindent	\textbf{Facial Codebook Pair.}
In general, VQ-VAE follows the variational auto-encoding diagram by first employing an encoder $\mathcal{E}$ to embed any facial expression $\mathbf{x}_t$ into a latent representation $\mathbf{z}_t \in \mathbb{R}^{h \times d}$: $\mathbf{z}_t = \mathcal{E}(\mathbf{x}_t)$, which consists of $h$ feature embeddings with the dimensionality of $d$. Differently, VQ-VAE involves a discrete codebook prior $\mathcal{C}=\{\mathbf{c}_k \in \mathbb{R}^{d}\}_{k=1}^{K}$ that allows any encoded $\mathbf{z}_t$ to be represented with a set of selected codebook tokens $\mathcal{S}$: $\{\mathbf{c}_k\}_{k\in\mathcal{S}}$. An element-wise quantization process is enforced to achieve this:
\begin{equation}
\label{eq:eq1}
\mathbf{q}_t = \mathrm{Qunt}(\mathbf{z}_t).
\end{equation}
The quantizer $\mathrm{Qunt}(\cdot)$ in Eq. \ref{eq:eq1} simply replaces every entry in the original $\mathbf{z}_t$ with its searched nearest neighbor from the entire codebook tokens, following:
\begin{equation}
\label{eq:eq2}
\mathrm{Qunt}(\mathbf{z}_t) = \underset{\mathbf{c}_k \in \mathcal{C}}{\operatorname{arg\, min}}\norm{\mathbf{z}_t - \mathbf{c}_k}.
\end{equation}
Given the quantized latent code $\mathbf{q}_t$,  VQ-VAE then re-produces the input in the motion space with a decoder $\mathcal{D}$ for self-reconstruction: $\hat{\mathbf{x}}_t = \mathcal{D}(\mathbf{q}_t)$.

We argue that pushing the entire facial data in one joint latent space is less effective due to the following two considerations: (i) different facial parts vary significantly in regard to the correlation with the speech audio. For example, lips exhibit stronger dependency on audio to accurately capture the corresponding sound, while upper faces are prone to be loosely correlated to speech but reflect more emotional variations. Eventually, this may lead the generated talking faces to static upper-face motions, even though the lips well follow the audio; (ii) as will be discussed in Sec. \ref{sec:cnt_syn}, a single latent space imposes challenges in partially controlling the facial movements. 

\begin{figure}[t]
\begin{center}
\includegraphics[width=0.95\linewidth]{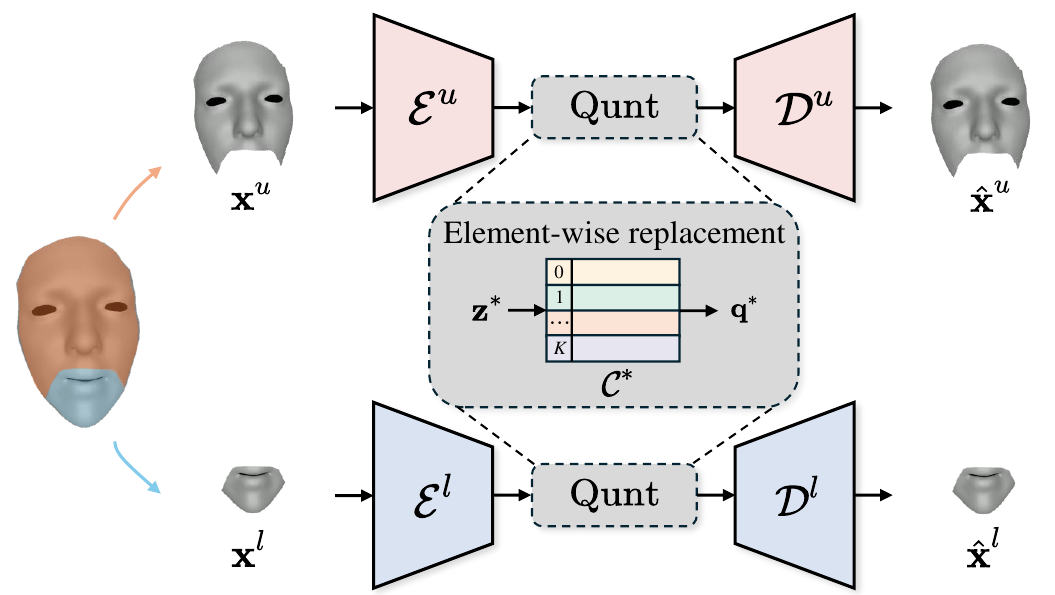}
\end{center}
\caption{\textbf{Codebook pair learning with VQ-VAEs} for lip (bottom) and upper-face areas (top).  $*\in\{u,l\}$ refers to upper-face or lip, respectively.}
\label{fig:codebook}
\end{figure}

We are thus motivated to learn individual priors for different facial parts. Specifically, the full-facial expression is dual-partitioned into lip $\mathbf{x}^l_t$ and upper-face $\mathbf{x}^u_t$ areas which further have their exclusive learnable modules and codebooks. As shown in Fig. \ref{fig:codebook}, we prepare a pair of (encoder, codebook, decoder) triplets   $(\mathcal{E}^l,\mathcal{C}^l,\mathcal{D}^l)$ and $(\mathcal{E}^u,\mathcal{C}^u,\mathcal{D}^u)$ to learn lip- and upper-face-codes $\mathbf{z}^l$ and $\mathbf{z}^u$, respectively. Both encoder-decoder pairs are constructed with self-attention. Such a strategy jointly mitigates the inherent correlation bias with audio between facial parts and contributes to improved diversity and controllability, thanks to the context-rich attribute of the quantized latent space.  

\noindent	\textbf{Training.} Each VQ-VAE is optimized with the following loss\footnote{Note that we omit the timestep for brevity without the loss generality}: 
\begin{equation}
\label{eq:eq3}
\begin{aligned}
\mathcal{L}^{*}_{vq} =& ||\mathbf{x}^{*}-\hat{\mathbf{x}}^{*}||^2 \\
+ & ||\mathrm{sg}(\mathbf{z}^{*})-\mathbf{q}^{*})||^2 + ||\mathbf{z}^{*}-\mathrm{sg}(\mathbf{q}^{*})||^2,
\end{aligned}
\end{equation}
in which $*\in\{u,l\}$ and $\mathrm{sg}(\cdot)$ denotes the stop-gradient operation introduced to combat the tendency of the non-differentiability of quantization. The first objective in Eq. \ref{eq:eq3} supervises motion self-reconstruction, while the latter two terms regularize the latent embeddings to approximate the discrete tokens such that the codebook can be well enriched.  We next need to determine how to exploit the facial prior pair for synthesis.

\subsection{Diverse Facial Motion Synthesis}
\label{sec:div_syn}
To achieve facial motion synthesis, our model temporally predicts the corresponding discrete latent code as a proxy of motion representation itself based on the audio input, in an auto-regressive manner.

\noindent	\textbf{Diverse Code Querying.} To encourage synthesis diversity, we propose to generate $N$ latent codes $\{\hat{\mathbf{z}}_t^{(n)}\}^{N}_{n=1}$ at every timestep for each input audio whose decoded motion representation $\{\hat{\mathbf{x}}_t^{(n)}\}^{N}_{n=1}$ can be richly diversified. To this end, we leverage a diversity-promoting objective
\begin{equation}
\label{eq:eq4}
\mathcal{L}_{d}(\{\hat{\mathbf{x}}_t^{(n)}\}^{N}_{n=1}) = -\sum^T_{t=1}\underset{i \neq j \in \lbrace1,\dots,N\rbrace}{\min}\norm{\hat{\mathbf{x}}^{{(i)}}_t-\hat{\mathbf{x}}^{{(j)}}_t}
\end{equation}
to yield duplication-aware diversification by penalizing the minimum pairwise sample distance along the entire temporal axis. Conceptually, Eq. \ref{eq:eq4} drives the model to cover diverse modes of the discrete latent space by identifying the ideal codes for optimization.

As we aim at diversity, forcing all of the synthesized motions to match the given single ground truth would induce conflict with Eq. \ref{eq:eq4}. We therefore modify the reconstruction loss to 
\begin{equation}
\label{eq:eq5}
\mathcal{L}_{rc}(\{\hat{\mathbf{x}}_t^{(n)}\}^{N}_{n=1}) = \sum^T_{t=1}\underset{i \in \lbrace1,\dots,N\rbrace}{\min}\norm{\mathbf{x}_t-\hat{\mathbf{x}}^{{(i)}}_t}
\end{equation}
such that at least one generated sample can hopefully characterize the ground truth.

%To nonetheless constrain all the samples and achieve realistic diversification, 

\begin{comment}
However,  Eq. \ref{eq:eq5} only conveys the supervision to one sample, leaving the remaining $N-1$ facial motions unconstrained. To nonetheless constrain all the samples, we are inspired by the observation that, although the dataset only provides one-to-one paired audio-motion data, one speech context is often recorded by multiple speakers. This means, while temporally unaligned, the dataset includes multiple facial motion data which share similar movement trends. For each ground-truth training audio-visual pair, we select $P$ facial motions with the same audio context to create the pseudo ground truth set $\{\mathbf{X}^{p}\}^P_{p=1}$. The resulting pseudo supervision is imposed by 
\begin{equation}
\label{eq:eq6}
%\mathcal{L}_{p}(\{\hat{\mathbf{X}}^{(n)}\}^{N}_{n=1}) = \sum_{p=1}^{P}\underset{i \in \lbrace1,\dots,N\rbrace}{\min}\mathrm{DTW}(\hat{\mathbf{X}}^{(i)}, \mathbf{X}^{p}),
\mathcal{L}_{p}(\{\hat{\mathbf{X}}^{(n)}\}^{N}_{n=1}) = \sum_{p=1}^{P}\mathrm{DTW}(\hat{\mathbf{X}}^{(i)}, \mathbf{X}^{p}),
\end{equation}
where $\mathrm{DTW}(\cdot,\cdot)$ performs soft Dynamic Time Warping \cite{cuturi2017soft} to handle the issues of varied sequence length and the dilatation due to differing personal speaking style. Intuitively, Eq. \ref{eq:eq6} provides weak supervision to guide each prediction to hopefully own similar temporal motion trends of the pseudo ground truths.
%\noindent	\textbf{Remark.} Although we can similarly let one of the 
\end{comment}

\begin{figure}[b]
\begin{center}
\includegraphics[width=1.0\linewidth]{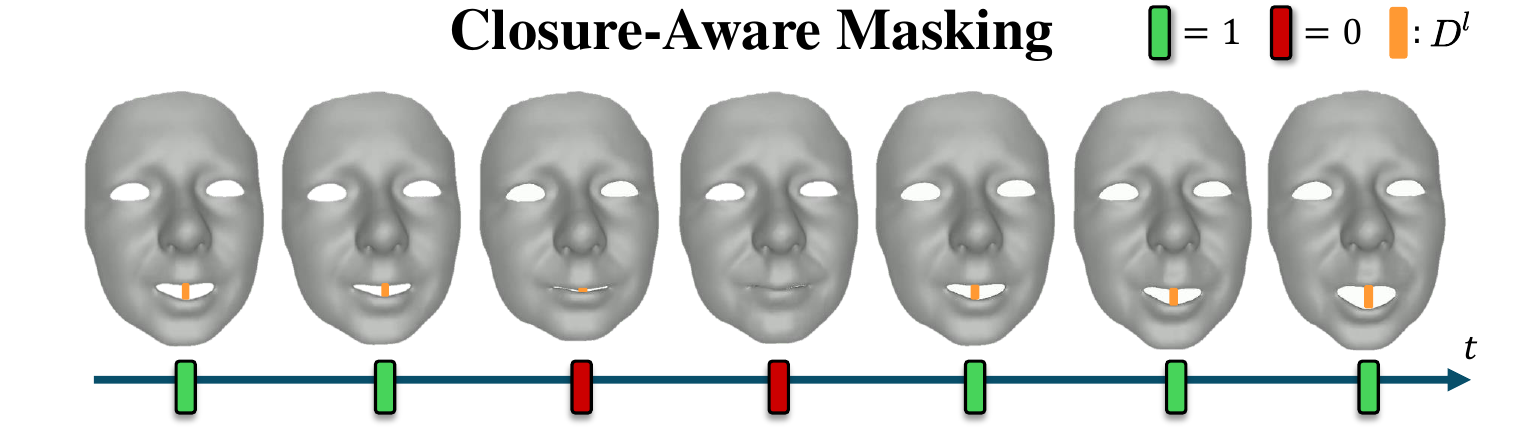}
\end{center}
\caption{\textbf{Illustration of Closure-aware masking.} The mask prevents the diversification from being promoted over the sounds with closed lip movements. }
\label{fig:closuremask}
\end{figure}

\noindent	\textbf{Closure-Aware Masking.}
Despite the reconstruction force in Eq. \ref{eq:eq5}, it only conveys the supervision to one sample, leaving the remaining $N-1$ facial motions unconstrained. More specifically, due to the diversification penalty imposed by Eq. \ref{eq:eq4}, the model can be forced to only pursue diversity by significantly sacrificing the audio fidelity. This issue can cause the generation to poorly characterize the plosive sounds in phonetics, such as ``b'' or ``p'', by unexpectedly extending the mouth shape for diversity. To mitigate this issue, we are inspired by the observation that, despite the diverse talking movements, humans always accurately close their mouths to pronounce the syllables that require the closure of both lips. This means the strength of the diversity loss should only be conveyed on those syllables that require sufficient mouth opening. 
For each training sequence, we prepare a binary mask sequence $\mathbf{M}=\{{m}_t\}_{t \in T}$  whose temporal entries are given by
\begin{equation}
\label{eq:eq6}
m_t = 
\begin{cases}
    1 & \text{if $D_t^l > \epsilon$} \\
    0 & \text{otherwise},
\end{cases}
\end{equation}
where $D_t^l$ measures the distance between the selected upper-lower-lip vertex pair at the $t$-th frame and $\epsilon$ is the pre-determined threshold. Fig. \ref{fig:closuremask} depicts our mask design. The diversification loss can be thus adapted to respect the mask during learning, which we will detail in Sec. \ref{sec:cnt_syn} the implementation.

In summary, our framework can be explained as a loss-driven diverse latent code query learning strategy without demanding the dataset to provide the required one-to-many supervision, yet also encouraging realism by enforcing masking guidance to respect the plosive syllables. We next discuss how to resolve controllability.

\begin{figure*}[t]
\begin{center}
\includegraphics[width=0.93\linewidth]{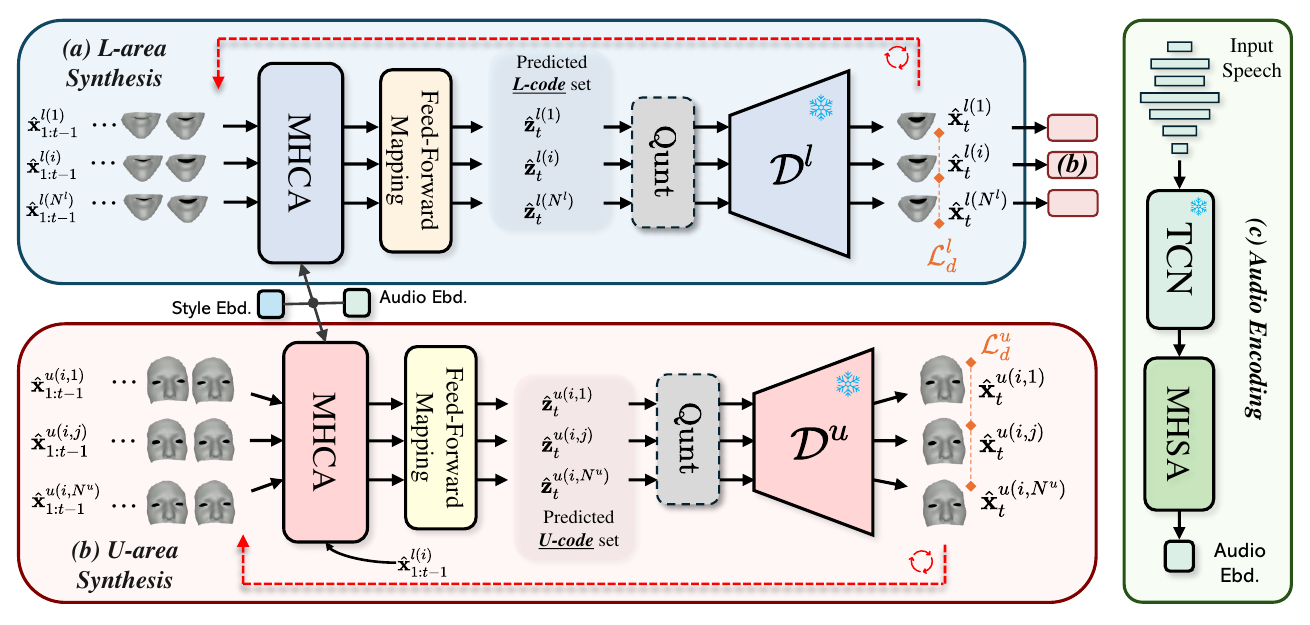}
\end{center}
\caption{\textbf{Method overview of CDFace.} Our method sequentially predicts diverse codes for the Lip- (L) and Upper-face (U)-areas, in \textit{(a)} and \textit{(b)} respectively, using the encoded audio embedding in \textit{(c)}. }
\label{fig:overview}
\end{figure*}

\subsection{Partial Controllable Synthesis}
\label{sec:cnt_syn}
Although our framework described above addresses synthesis diversity, it does not straightforwardly provide any control over specific facial parts. To tackle this problem, we design our model to sequentially query diverse codes for different facial parts, as illustrated in Fig. \ref{fig:overview}.

\noindent	\textbf{Sequential Modeling for Facial Parts.} Following the discussion in Sec. \ref{sec:vq-vaes}, our model starts with the prediction of the lip (L)-code set. Specifically, the input speech $\mathbf{A}$ is first embedded with the audio encoder $\mathcal{E}^a$ to produce the audio feature $\mathbf{F}^a$: $\mathbf{F}^a = \mathcal{E}^a(\mathbf{A})$. We adopt the strategy in \cite{xing2023codetalker,fan2022faceformer} by using the trained wav2vec 2.0 \cite{baevski2020wav2vec} model which involves a temporal convolutions network (TCN) followed by a self-attention module to structure $\mathcal{E}^a$. Then, based on the produced past lip predictions $\{\hat{\mathbf{x}}_{1:t-1}^{l(i)}\}_{i=1}^{N^l}$, the ``new'' set of $N^l$ L-codes  $\{\hat{\mathbf{z}}_t^{l(i)}\}_{i=1}^{N^l}$  are generated in an auto-regressive manner. We also include a learnable style token $\mathbf{s}$ as in \cite{xing2023codetalker,fan2022faceformer} during learning. The resulting temporal code querying module is devised with multi-head cross attention (MHCA) considering the domain discrepancy between audio and motion, followed by a feed-forward (FF) mapping to project the output in the latent space. Specifically, we use a $N^l$-head FF network to output the code set. Such a recursive process can be thus formalized as
\begin{equation}
\label{eq:eq7}
\{\hat{\mathbf{z}}_t^{l(i)}\}_{i=1}^{N^l} =  \text{MLP}(\text{MHCA}(\{\hat{\mathbf{x}}_{1:t-1}^{l(i)}\}_{i=1}^{N^l}, \mathbf{F}^a, \mathbf{s})),
\end{equation}
where $\{\hat{\mathbf{z}}_t^{l(i)}\}_{i=1}^{N^l}$ can be easily decoded into their lip motions.

For \textit{each} L-code $\hat{\mathbf{z}}_t^{l(i)}$, we further predict $N^u$ upper-face (U)-codes $\{\hat{\mathbf{z}}_t^{u(i,j)}\}_{j=1}^{N^u}$ to eventually form multiple full-face motions. Here, $\hat{\mathbf{z}}_t^{u(i,j)}$ denotes the $j$-th U-code paring the $i$-th L-code. In addition to the past face sequence $\{\hat{\mathbf{x}}_{1:t-1}^{u(i,j)}\}_{j=1}^{N^u}$ and audio embedding $\mathbf{F}^a$, we include the past lip motion $\hat{\mathbf{x}}_{1:t-1}^{l(i)}$ to improve the inter-parts coherence during the ``new'' U-code set generation, using MHCA:
\begin{equation}
\label{eq:eq8}
\{\hat{\mathbf{z}}_t^{u(i,j)}\}_{j=1}^{N^u} = \text{MLP}(\text{MHCA}(\{\hat{\mathbf{x}}_{1:t-1}^{u(i,j)}\}_{j=1}^{N^u}, \hat{\mathbf{x}}_{1:t-1}^{l(i)}, \mathbf{F}^a, \mathbf{s})).
\end{equation}

\noindent	\textbf{Training.} We here give our new training objective for each facial portion. The upper-face region simply adopts
\begin{eqnarray}
\label{eq:eq9}
\mathcal{L}^u_{d}=\sum_i^{N^l}\mathcal{L}_{d}(\{\hat{\mathbf{x}}_t^{u(i,j)}\}_{j=1}^{N^u})\\
%\mathcal{L}^l_{d}=\mathcal{L}_{d}(\{\hat{\mathbf{x} }^{l(i)}_t \cdot m_t\}_{i=1}^{N^l}), 
\label{eq:eq10}
\mathcal{L}^u_{rc}=\sum_i^{N^l}\mathcal{L}_{rc}(\{\hat{\mathbf{x}}_t^{u(i,j)}\}_{j=1}^{N^u}).
\end{eqnarray}
to ensure diversity and reconstruction, while for the lip region, we compute the diversity loss with the maksed prediction: 
\begin{equation}
\label{eq:eq11}
\mathcal{L}^l_{d}=\mathcal{L}_{d}(\{\hat{\mathbf{x} }^{l(i)}_t \cdot m_t\}_{i=1}^{N^l}), 
\end{equation}
to promote closure-aware diversification. Moreover, in terms of the lip reconstruction, we further include the supervision over the sound with closure lip movements in all predictions using ground truth. The modified lip reconstruction loss is given by 
\begin{equation}
\label{eq:eq12}
\mathcal{L}^l_{rc}=\mathcal{L}_{rc}(\{\hat{\mathbf{x} }^{l(i)}_t\}_{i=1}^{N^l}) + \frac{1}{N^l}\sum^{N^l}_{i=1}\sum^T_{t=1}\norm{(\mathbf{x}^l_t-\hat{\mathbf{x}}^{{l(i)}}_t)(1-m_t)}. 
\end{equation}
In addition, we apply feature-level regularizers for each facial portion to let the predicted codes stay within the corresponding codebook
\begin{eqnarray}
\label{eq:eq13-14}
\mathcal{L}^l_{rg} = \sum_{i=1}^{N^l}\sum_{t=1}^{T}\norm{\hat{\mathbf{z}}^{l(i)}_t-\mathrm{sg}(\mathbf{q}^{l(i)}_t)},\\
\mathcal{L}^u_{rg} = \sum_{i=1}^{N^l}\sum_{j=1}^{N^u}\sum_{t=1}^{T}\norm{\hat{\mathbf{z}}^{u(i,j)}_t-\mathrm{sg}(\mathbf{q}^{u(i,j)}_t)}.
\end{eqnarray}
The final training losses we aim to optimize for each facial part can be expressed as
\begin{eqnarray}
\label{eq:eq15}
\small
\mathcal{L}^l = \lambda_{d}^{l}\mathcal{L}^l_{d} +  \lambda^l_{rc}\mathcal{L}^l_{rc}+ \lambda_{rg}\mathcal{L}^l_{rg},\\
\label{eq:eq16}
\mathcal{L}^u = \lambda_{d}^{u}\mathcal{L}^u_{d} + \lambda^u_{rc}\mathcal{L}^u_{rc} + \lambda_{rg}\mathcal{L}^u_{rg}.
\end{eqnarray}
$(\lambda_{d}^{l}, \lambda_{d}^{u}, \lambda^l_{rc}, \lambda^u_{rc}, \lambda_{rg})$ denote the weights to control the strength of each term. In particular, each part of our model (i.e., (a) and (b) in Fig. \ref{fig:overview}) can be separately trained with Eq. \ref{eq:eq15} or \ref{eq:eq16}, or end-to-end optimized by combining these two losses. Our method contributes to diversity and controllability in a unified formulation. Once trained, one can strictly control one part by fixing the latent codes while varying those for the other part for diversification.

\section{Experiment}
In this section, we conduct a series of experiments to evaluate the effectiveness of our model against other state-of-the-art speech-driven facial animation methods. 

\noindent	\textbf{Dataset.} Following \cite{fan2022faceformer,xing2023codetalker,stan2023facediffuser}, we evaluate on two widely employed vertex-based talking face datasets: BIWI \cite{fanelli20103} and VOCASET \cite{cudeiro2019capture}. 

\textbf{BIWI \cite{fanelli20103}} is originally collected to investigate affective talking states with 4D facial scans. It comprises in total 40 sentences spoken by 14 human subjects where six males and eight females are involved. All speakers are directed to repeat the same sentence twice with and without emotional tones during recording. The average sentence length is 4.67 seconds. The meshes are captured to reflect dense facial geometries with 23370 vertices at 25Hz. For fair comparisons, we follow the data split in \cite{fan2022faceformer, xing2023codetalker} to use the BIWI-Train that contains 192 sentences and BIWI-Val with 24 sentences, both from 6  subjects. The testing sets have two parts: BIWI-Test-A and BIWI-Test-B. For BIWI-Test-A, it contains 24 sentences by six seen subjects, which can be thus utilized for both quantitative and qualitative evaluations. In regard to BIWI-Test-B, it includes 32 sentences with eight unseen subjects and is only used for qualitative understanding.

\textbf{VOCASET \cite{cudeiro2019capture}} consists of 480 facial motions with 12 subjects. It records them at 60Hz, with each sentence being approximately 4 seconds long. All facial meshes follow the FLAME \cite{li2017learning} topology registration to
have 5023 vertices. To be consistent with \cite{fan2022faceformer, xing2023codetalker}, we adopt the split in \cite{cudeiro2019capture} to create VOCA-Train, VOCA-Val, and VOCA-Test, for training, validation, and testing, respectively. As VOCASET only contains unseen testing subjects during training, we follow \cite{xing2023codetalker,stan2023facediffuser,fan2022faceformer} by only performing qualitative evaluation on it.

\noindent	\textbf{Implementation Details.} The training comprises VQ-VAEs and the sequential facial code querying model (CDFace). We individually train each VQ-VAE for the lip and upper face for 200 epochs on both datasets. For CDFace, we further train each part for 100 and 50 epochs on BIWI and VOCASET, respectively, where the corresponding decoder VQ-VAE for each facial part is kept frozen. This is mainly to relax the  GPU limitation considering the high dimensionality of 3D meshes. Inspired by \cite{xing2023codetalker,fan2022faceformer}, we enforce teacher-forcing during training while following the auto-regressive manner of synthesis in inference. We set $(\lambda_{d}^{l}, \lambda_{d}^{u}, \lambda^l_{rc}, \lambda^u_{rc}, \lambda_{rg}, \epsilon)$ to $(0.2, 0.2, 10, 10, 20, 0.01)$ for BIWI and $(0.02, 0.02, 1, 1, 1, 0.005)$ for VOCASET. All  the training adopts the AdamW \cite{loshchilov2017decoupled} optimizer.

\subsection{Quantitative Evaluation}
\label{sec:quantitative_eva}
We first report the quantitative evaluation results against prior state-of-the-art speech-driven facial animation synthesis methods.
Specifically, for deterministic methods, we compare against
%VOCA \cite{cudeiro2019capture}, MeshTalk \cite{richard2021meshtalk}, 
FaceFormer \cite{fan2022faceformer} and CodeTalker \cite{xing2023codetalker}, while for stochastic models, 
we compare with FaceDiffuser \cite{stan2023facediffuser}. 
Since the deterministic models are inherently different from stochastic ones, a straightforward comparison against these methods would be less feasible. To nonetheless ensure a fair comparison, we devise a \textit{deterministic version} of our model by simply setting ($N^l, N^p$) to ($1,1$) and modifying the weights for diversification $(\lambda_{d}^{l}, \lambda_{d}^{u})$ to ($0,0$) to retrain our model.

\noindent	\textbf{Evaluation Metrics.}  We separate the evaluation metrics for deterministic and stochastic cases. For deterministic scenarios, we follow \cite{xing2023codetalker,stan2023facediffuser} by adopting the following metrics:
\begin{itemize}
    \item Lip vertex error (LVE). LVE measures the deviation of the generated lip vertices relative to the ground truth, which is derived by computing the frame-wise maximal $\mathcal{L}$2 error and then averaging over all frames.
    \item Mean vertex error (MVE). MVE is similar to the LVE metric but extends the calculation for averaged vertex error to the whole facial region.    
    \item Upper-Face Dynamics Deviation (FDD). FFD calculates the deviation of the generated upper-face vertices with respect to the ground truth. Specifically, given the predicted $\hat{\mathbf{X}}$ and the ground-truth $\mathbf{X}$ facial motions, FDD is formalized as 
    \begin{equation}
    \label{eq:eq17}
    \text{FDD}(\hat{\mathbf{X}}, \mathbf{X}) = \frac{1}{V^u}({\text{std}(\hat{\mathbf{X}}^u)-\text{std}(\mathbf{X}^u)}), 
    \end{equation}
    where $\text{std}(\cdot)$ calculates the standard deviation 
    of the $\mathcal{L}$2 distance for each vertex at all timesteps, and $V^u$ denotes the number of upper-face vertices. FFD indicates how close the upper face moving trend is compared to the ground truth. 
\end{itemize}

\begin{table}[t]
\caption{\textbf{Quantitative evaluation of deterministic prediction on BIWI-Test-A}. The best and the second-best results are highlighted in bold and underlined, respectively.}
\centering
\label{table1}
\scalebox{1.1}{
\begin{tabular}{lccc}
\toprule
\multicolumn{1}{c}{} & \specialcell{ LVE $\downarrow$\\ ($\times 10^{-4}$mm)}   & \specialcell{ FDD $\downarrow$\\ ($\times 10^{-5}$mm)}    & \specialcell{ MVE $\downarrow$\\ ($\times 10^{-4}$mm)}   \\ \hline
FaceFormer    \cite{fan2022faceformer}       & 5.610 & 4.732 & 10.732 \\
CodeTalker  \cite{xing2023codetalker}         & 4.777 & 4.111  & 7.576 \\
FaceDiffuser \cite{stan2023facediffuser}        & \textbf{4.282} & \underline{4.042}  & \textbf{6.885} \\ \hdashline
\textit{CDFace}                 & \underline{4.498} & \textbf{3.231}  & \underline{7.572} \\ \bottomrule
\end{tabular}}
\end{table}

\begin{table}[t]
\caption{\textbf{Quantitative evaluation of stochastic synthesis on BIWI-Test-A}. The best results are highlighted in bold.}
\centering
\label{table2}
\scalebox{1.05}{
\begin{tabular}{lcccc}
\toprule
             & \specialcell{ APD $\uparrow$\\ (mm)}    & \specialcell{ UPD $\uparrow$\\ (mm)}   & \specialcell{ LPD $\uparrow$\\ (mm)}   & \specialcell{ MPD $\uparrow$\\ (mm)}    \\ \hline
FaceDiffuser \cite{stan2023facediffuser} & 2.423$e^{-3}$   & 1.256$e^{-3}$ & 9.895$e^{-4}$  & 2.209$e^{-3}$   \\
\textit{CDFace}       & \textbf{12.180} & \textbf{7.850} & \textbf{4.167} & \textbf{10.510} \\ \bottomrule
\end{tabular}}
\end{table}

\begin{table}[t]
\caption{\textbf{Quantitative evaluation of controllable synthesis on BIWI-Test-A} against FaceDiffuser. RS denotes rejection sampling for better control. }
\centering
\label{table3}
\scalebox{1.2}{
\begin{tabular}{lcccc}
\toprule
             &  &\specialcell{ UPD $\uparrow$\\ (mm)}      & \specialcell{ LPD $\downarrow$\\ (mm)}      &  \\\hline
FaceDiffuser (w. RS) &  & 1.237$e^{-3}$ & 9.834$e^{-4}$ &  \\
\textit{CDFace}      &  & \textbf{7.850}    & \textbf{0.0}        &  \\ \bottomrule
\end{tabular}}
\end{table}

For stochastic predictions, we compare the diversity regarding 
\begin{itemize}
    \item Average Pairwise Distance (APD). We assess the per-speech motion diversity with APD, following
    \begin{equation}
    \label{eq:eq18}
    \text{APD}(\{\hat{\mathbf{X}}^{(i)}\}^S_{i=1}) = \frac{1}{S(S-1)}\sum_{i=1}^S \sum_{j=1, j\neq i}^S||\hat{\mathbf{X}}^{(i)}-\hat{\mathbf{X}}^{(j)}||,
    \end{equation}
    where $S$ is the sample number. APD computes the average $\mathcal{L}$2 distance between all synthesized talking face pairs to investigate diversity. 
    \item Upper-Face/Lip Pairwise Distance (UPD/LPD). UPD and LPD are identical to the APD calculation but individually assess the motion diversity of the upper-face and lip regions. We set $S$ to five for all the comparisons of stochastic synthesis.
    \item Minimum Pairwise Distance (MPD). We further introduce MPD to measure the similarity for the closest two results among all the generation pairs
    \begin{equation}
    \label{eq:eq19}
    \text{MPD}(\{\hat{\mathbf{X}}^{(i)}\}^S_{i=1}) = 
    {\min}_{i \neq j \in \lbrace1,\dots,S\rbrace}\norm{\hat{\mathbf{X}}^{(i)}-\hat{\mathbf{X}}^{(j)}}.
    \end{equation}
    A lower MPD indicates a higher sample resemblance. 
\end{itemize}

\begin{figure*}[t]
\begin{center}
\includegraphics[width=1.0\linewidth]{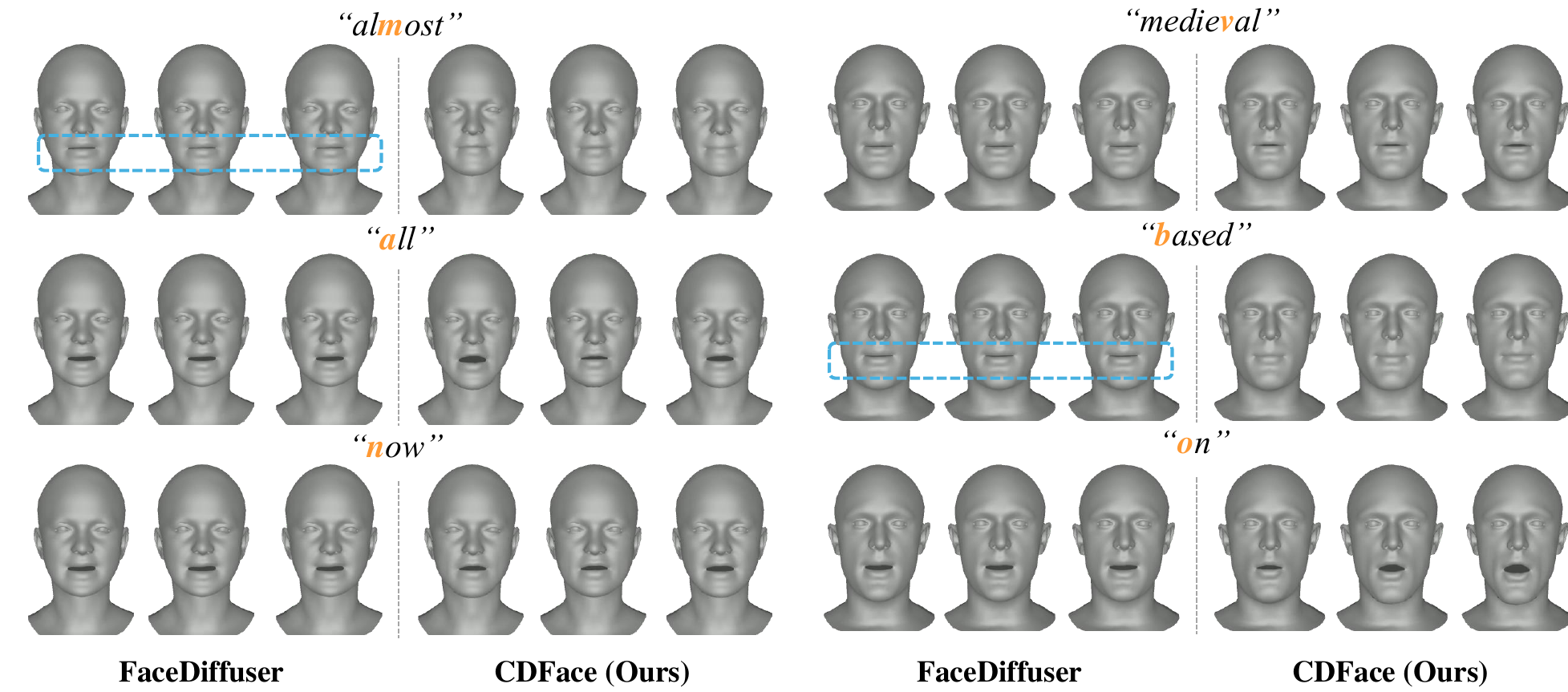}
\end{center}
\caption{\textbf{Diverse synthesis on VOCASET-Test} against FaceFormer. For each syllable, we display three samples from CDFace and FaceDiffuser, respectively.}
\label{fig:qual_voca}
\end{figure*}

\begin{figure*}[t]
\begin{center}
\includegraphics[width=1.0\linewidth]{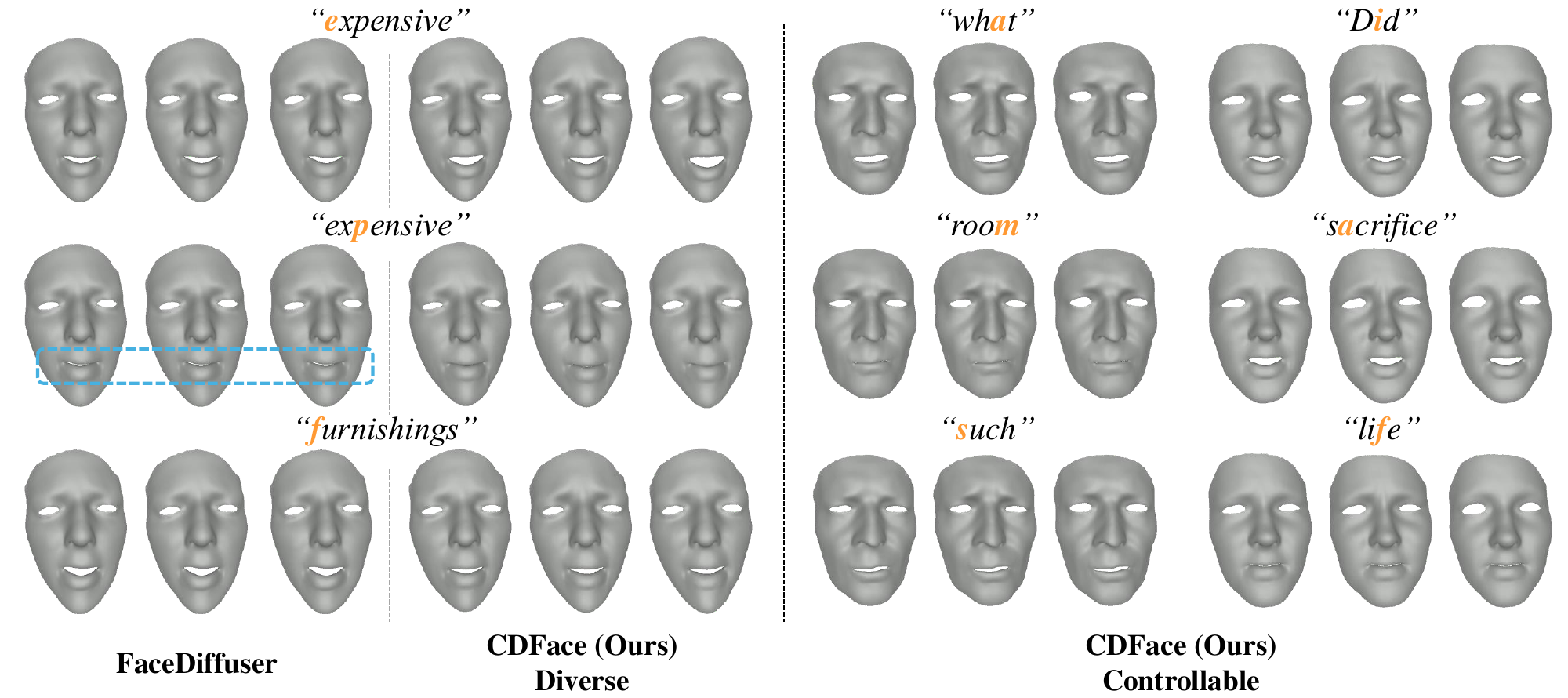}
\end{center}
\caption{\textbf{Diverse (left) and controllable (right) synthesis on BIWI-Test-B} against FaceFormer. For each syllable, we display three samples from the corresponding method, respectively.}
\label{fig:qual_biwi}
\end{figure*}

\begin{figure}[t]
\begin{center}
\includegraphics[width=0.49\linewidth]{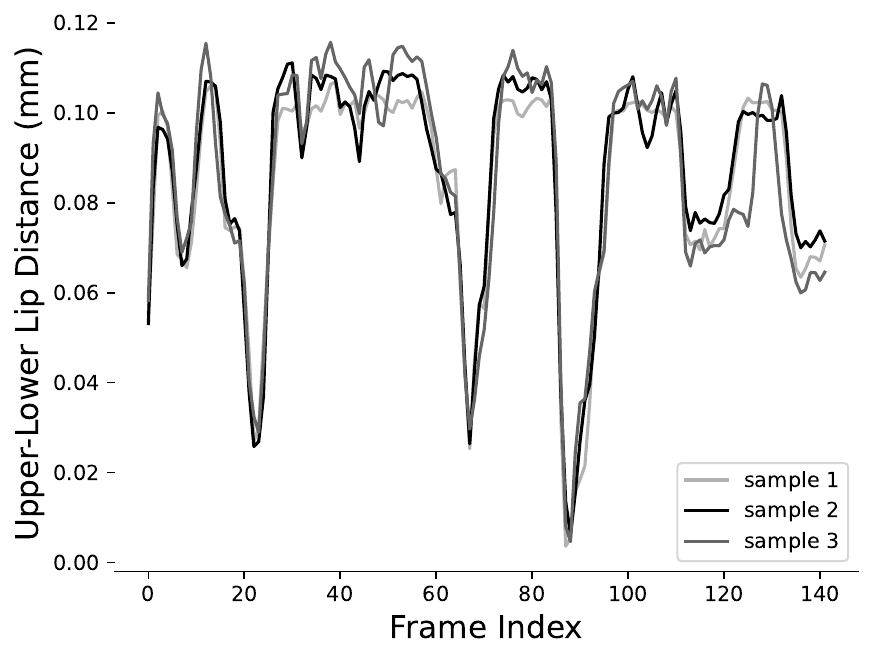}
\includegraphics[width=0.49\linewidth]{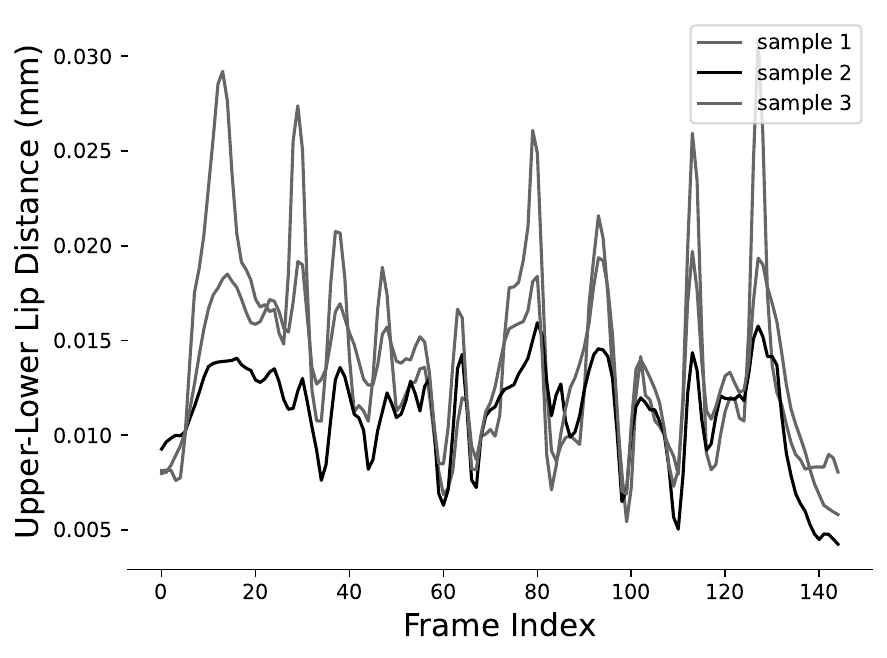}
\end{center}
\caption{\textbf{Changes of the upper-lower lip distance} of three examples produced by CDFace on BIWI-Test-A (left) and VOCASET (right). 
}
\label{fig:lip_curve}
\end{figure}

The results are summarized in Tabs. \ref{table1} and \ref{table2}. It can be observed from Tab. \ref{table1} that CDFace achieves comparable lip synchronization performance to the compared approaches in its deterministic mode. For the FDD metric, CDFace outperforms all other state of the arts, indicating a better characterization of upper facial expression motion trends. This is because since the facial priors are separately prepared for upper-face and lip regions, the inconsistent dependencies of different facial parts on the audio can be individually modeled to alleviate the mapping ambiguity. The fact that our method outperforms CodeTalker \cite{xing2023codetalker}, in which only one prior for all facial parts are involved, further indicates the effectiveness of region-wise prior modeling. 

For stochastic synthesis, we can see from Tab. \ref{table2}
that our method outperforms FaceDiffuser \cite{stan2023facediffuser} by a large margin regarding sample diversity in all metrics, including both upper-face (UPD) and lip (LPD) regions. The primary reason is that, the simple diffusion-based generative modeling can hinder the coverage of the entire modality of the data distribution with likelihood-based sampling, which causes FaceDiffuser to suffer from severe mode collapse. By contrast, since CDFace is designed to employ a diversity-promoting loss, it is forced to entirely explore different data modes to diversify the results. Besides, a high MPD indicates that the generated samples are diversified without duplicated pairs. 

To further quantitatively investigate the validity of the samples produced by CDFace, we plot in Fig. \ref{fig:lip_curve} the change of the upper-lower lip distance of three samples along the temporal axis. We notice that the mouth amplitudes tend to vary more drastically on VOCASET compared to those on BIWI. On both datasets, the samples follow a consistent moving trend where the opening and closing moments are generally synchronized, particularly regarding the motion transition between closing and opening sounds. This confirms the capacity of our model to synthesize multiple talking samples with high audio fidelity.

We next quantitatively assess the performance of controllable synthesis against FaceFormer in Tab. \ref{table3}. Here, the synthesis is aimed to produce the \textbf{\textit{same}} lip motion but \textbf{\textit{diverse}} upper-face motions, which is less studied in prior arts. In particular, we adapt rejection sampling (RS) to FaceFormer for better control of the lip region, where we sample 30 facial motions from the diffusion model and select five with lip motions closest to the target one. We can see that RS does not contribute noticeably to the performance of FaceDiffuser, which we assume to be mostly due to the mode collapse issue. Anyway, compared to Tab. \ref{table2}, the controllability is slightly improved with RS by sacrificing some diversity for the upper face. On the contrary, since CDFace naturally provides partial motion control due to the sequential design, it jointly pursues high diversity and controllability for different facial parts.

\subsection{Qualitative Evaluation}
We here report the qualitative results of our method. As our method is designed to synthesize stochastic talking faces, we compare the results against the diffusion-based model, FaceFormer, for visual understanding. The results on VOCASET and BIWI are presented in Figs. \ref{fig:qual_voca} and \ref{fig:qual_biwi}, respectively, where three samples are visualized for each method.

\noindent	\textbf{Diverse Synthesis.} It can be observed in the blue dotted area that on both datasets, FaceDiffuser cannot accurately characterize the closing movements for the lips regarding the syllables that require mouth closure. Also, the non-deterministic samples produced by FaceDiffuser share a significant visual resemblance, which is also reflected in the low diversity metrics in Tab. \ref{table2}. Similar to the analysis in Sec. \ref{sec:quantitative_eva}, as BIWI and VOCASET only have a limited number and variation of facial samples, conventional generative modeling can easily trigger 
mode collapse to deprive sample diversity on small datasets, thus causing the generation to be almost deterministic. By contrast, it can be confirmed that our method presents highly different movements, especially for the syllables allowing for potentially different pronunciation patterns, such as ``\textbf{\textit{o}}n'' (Fig. \ref{fig:qual_voca}, 3rd row, right) or ``\textbf{\textit{e}}xpensive'' (Fig. \ref{fig:qual_biwi}, 1st row, left). Moreover, for plosive sounds requiring mouth closure, our method well characterizes these syllables with precise lip-closing movements in all samples, which evidences the strength of CDFace in selectively diversifying the talking movements while maintaining high audio fidelity and realism. \textit{Please refer to the supplementary video for a clear visualization inspection.}

\begin{table}[]
\caption{\textbf{User study} statistics on BIWI-Test-B and VOCA-Test. We show the results of A/B and scoring tests on the top and bottom, respectively. }
\centering
\label{table_us}
\scalebox{1.2}{
\begin{tabular}{lccc}
\toprule
                      & \multicolumn{3}{c}{BIWI-Test-B} \\
\multicolumn{1}{c}{}  & Lip Sync  & Realism & Diversity \\ \hline
Ours vs. FaceFormer   & 52.90     & 55.75   & -         \\
Ours vs. CodeTalker   & 46.18     & 47.10   & -         \\
Ours vs. FaceDiffuser & 51.90     & 63.45   & 87.50     \\ \bottomrule
                      & \multicolumn{3}{c}{VOCASET}     \\
                      & Lip Sync  & Realism & Diversity \\ \hline
Ours vs. FaceFormer   & 63.45     & 61.53   & -         \\
Ours vs. CodeTalker   & 50.95     & 48.08   & -         \\
Ours vs. FaceDiffuser & 53.85     & 57.70   & 71.23     \\ \bottomrule
\end{tabular}}

\centering
\scalebox{1.05}{
\begin{tabular}{lcccc}
\\
\toprule
            & Lip Sync & Realism & Diversity & Expressiveness \\ \hline
BIWI-Test-B & 3.83     & 3.85    & 3.69      & 3.66           \\
VOCA-Test   & 4.28     & 4.19    & 4.34      & 3.83           \\ \bottomrule
\end{tabular}}

\end{table}

\noindent	\textbf{Controllable Synthesis.} We also provide in Fig. \ref{fig:qual_biwi}(right) the results for controllable synthesis. Our model yields strictly controlled lip movements with high diversity for the upper face. As the upper-face motions are loosely restrained by the audio compared to the lips, it reflects more emotional variations to interpret the given speech context. Specifically, diversification is primarily expressed in varied shapes for the eyes or frowned movements of the eyebrows. \textit{The animated results are included in the supplementary video, where we also present the results for controllable synthesis on VOCASET.}

%our method accurately characterizes the  plosive sound with lip closure for all the samples, 

\subsection{User Study}
Since the human visual system is still the most reliable measure in evaluating talking realism, we conduct a user study to perceptually assess the generation quality. Following \cite{fan2022faceformer,stan2023facediffuser,xing2023codetalker}, we adopt the A/B testing for each comparison against prior arts. In particular, we randomly sample one talking sample from the results produced by our method in a side-to-side manner to compare against the deterministic model (i.e., FaceFormer \cite{fan2022faceformer} and CodeTalker \cite{xing2023codetalker}), while in comparing against FaceDiffuser \cite{stan2023facediffuser}, we randomly sample two sequences per speech, and ask participants to select one talking group that performs better. The participants are required to judge lip synchronization, realism, and diversity (only for FaceDiffuser). We prepare overall 24 audio clips to generate talking faces, and eventually, 26 participants are involved in the evaluation. The results of the perceptual study are summarized in Tab. \ref{table_us}. Based on the feedback from the top table in Tab. \ref{table_us}, despite the randomly selected sample, our method outperforms FaceFormer and receives comparable positive feedback with CodeTalker on both datasets. Also, we find that while our method generally produces visually competitive samples with FaceDiffuser, it yields significantly higher talking diversity on both datasets. 

To better investigate the quality of diverse talking samples, we further present three samples per audio, and ask the participants to take a scoring test for our method. Specifically, participants are required to judge lip synchronization, realism, diversity, and expressiveness for each group of samples, and then rate on a scale of 1-5 (5 for the best). For example, as for the realism metric, 5 should be rated when one regards all three samples to be realistic, and 1 refers to that none of these samples seem realistic. We follow \cite{lin2024glditalker,lin2024emoface} by counting the mean opinion score (MOS), and tabulate the results in Tab. \ref{table_us}(bottom). It can be observed that our method receives high MOS in all metrics. We notice that the results on VOCASET generally achieve higher scores than BIWI. We expect this to be that, while VOCASET includes less upper-face variation, the lip motions are more expressive than BIWI, which leads to an easier configuration for balancing diversity and realism during optimization.

Based on the above A/B and scoring questionnaire study, we can confirm that our method produces both diverse and natural talking facial motions that are perceptually consistent with the audio.

\begin{figure}[t]
\begin{center}
\includegraphics[width=1.0\linewidth]{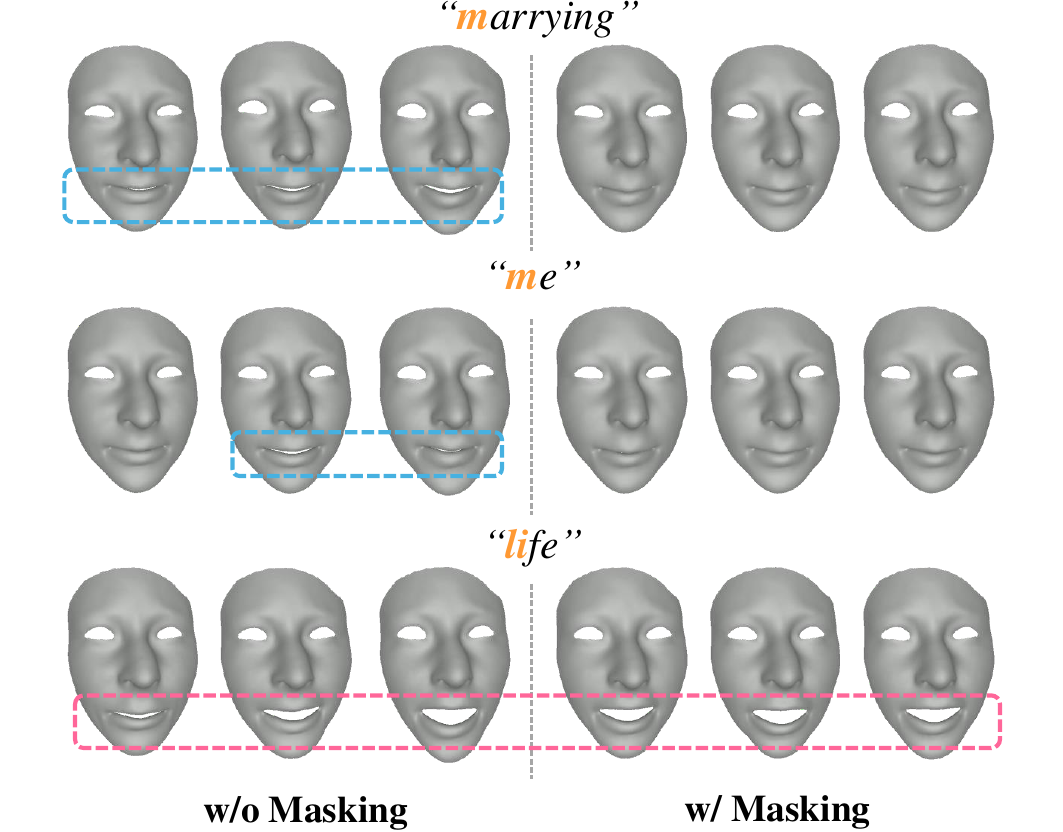}
\end{center}
\caption{\textbf{Qualitative results of ablation study for closure-aware masking} on BIWI.  
}
\label{fig:masking_ablation}
\end{figure}

\subsection{Ablation Studies}
To gain deeper insights into our method, we report the results of ablative evaluations to study several key components in our model. 

\noindent	\textbf{Closure-Aware Masking.} We manage to achieve closure-aware diversification by introducing a masking guidance for the lip area. To evaluate the influence, we here qualitatively study the results in Fig. \ref{fig:masking_ablation} for visual inspection. It can be seen that our model without the masking is more likely to predict lip movements that do not respect the syllables with plosive sounds (blue dotted area), while the results with our masking appear completely closed motions. This suggests the significance of the mask during training to realistic talking dynamics in predicting stochastic speech-driven facial motions. Also, we notice that there is in general a trade-off between audio fidelity and diversity. Despite the realism provided by the masking, we notice that it somehow sacrifices some diversity when removing it (magenta dotted area).

\noindent	\textbf{Number of Sample.} CDFace involves the number of facial motion samples as hyperparameters during training to produce differing patterns. To study the influence, we study the lip sync by varying $N^l$ and provide the change of diversity and realism in Tab. \ref{table4}. In assessing realism, we average the vertex error between all synthesized samples and the one ground truth. \textit{Note that this is, however, not the most proper manner for accurately evaluating the realism as different samples cannot possibly match the sole ground truth, and the comparison is mostly for reference to study the moving trend of facial samples.} We report Average LVE (ALVE) and LPD, respectively. We observe that a larger $N^l$ tends to yield higher diversity while sacrificing accuracy, which again, confirms the diversity-fidelity trade-off.

\begin{table}[t]
\caption{\textbf{Influence of the lip sample number} during training  on BIWI. }
\centering
\label{table4}
\scalebox{1.2}{
\begin{tabular}{lcc}
\toprule
     & \specialcell{ LPD $\uparrow$\\ (mm)}    &  \specialcell{ ALVE $\downarrow$\\ ($\times 10^{-3}$mm)}        \\ \hline
$N^l$=5  & 4.167 & \textbf{1.133}   \\
$N^l$=10 & 5.864 & 1.307 \\
$N^l$=15 & \textbf{7.715} & 1.308 \\ \bottomrule
\end{tabular}}
\end{table}

\begin{table}[t]
\caption{\textbf{Influence of the generation order} during training on BIWI. ``U'', ``L'' denote upper-face and lip region, respectively. }
\centering
\label{table5}
\scalebox{1.1}{
\begin{tabular}{lccc}
\toprule
\multicolumn{1}{c}{} & \specialcell{ LVE $\downarrow$\\ ($\times 10^{-4}$mm)}   & \specialcell{ FDD $\downarrow$\\ ($\times 10^{-5}$mm)}    & \specialcell{ MVE $\downarrow$\\ ($\times 10^{-4}$mm)}   \\ \hline
U $\rightarrow$ L                  & 4.597 & 3.466  & 7.614 \\ 
L $\rightarrow$ U                 & \textbf{4.498} & \textbf{3.231}  & \textbf{7.572} \\ \bottomrule
\end{tabular}}
\end{table}

\begin{comment}
\begin{figure}[t]
\begin{center}
\includegraphics[width=1.0\linewidth]{figure/limitation.pdf}

\end{center}
\caption{\textbf{A failure example} of the unnatural height of the right eye (right) produced in our results compared to the expressions with natural both-eye height (left). }
\label{fig:limitation}
\end{figure}
\end{comment}

\noindent	\textbf{Generation Order.} CDFace follows a pre-determined generation order from lip to upper face for compositional facial motion synthesis. To investigate the influence of such an order, we quantitatively examine the prediction accuracy in Tab. \ref{table5}, where we adopt the \textit{deterministic version} of our model. It can be found that enforcing the order from lip to upper yields increased accuracy. As the lip receives a stronger impact from the audio, introducing the upper face for lip prediction amplifies the mapping ambiguity for identical pronunciations to lower the prediction accuracy. Consequently, the generation ordered in lip to upper face contributes to better deterministic performance. 

\subsection{Limitations}
Despite the effectiveness, our method also involves some limitations that require improvement. We notice that, in diversifying the movements in VOCASET, CDFace can sometimes trigger overly rapid eyeball motions. Since the eye-region motions inherently contain less variation in VOCASET, which does not include blinking, achieving plausible yet differing upper-face motions for VOCASET can be challenging. Nonetheless, introducing a motion prior to the eye region can be expected to regularize such eyeball jitters. In addition, humans speak not only using their lips but also using a combination of tongues and teeth, which jointly contributes to diverse talking motions. However, since the currently released datasets rarely include such inner mouth parts of representations, CDFace also focuses on the modeling of general face shapes, like prior arts. Hence, experimenting on a dataset that compromises the entire facial components would also constitute an interesting future direction.

%We adopt the exact same dataset split for BIWI as done in [17, 54] and only use the emotional sequences subset. The split results into training set BIWITrain that contains 192 sentences and validation set BIWI-Val that contains 24 sentences from 6 training subjects. There are two test sets: BIWI-Test-A, containing 24 sentences from seen subjects and BIWI-Test-B, containing 32 sentences from the 8 remaining unseen subjects. The former test set is used for computing objective metrics while both facilitate the qualitative analysis and the perception study. BIWI-Test-B is further used to compute the diversity metric that we define in the next section. W

%\noindent	\textbf{Implementation Details.} The training involves three models: AinB-VAE, MDM, and diversity sampler (DS). For AinB-VAE, we train it for 500 epochs with a learning rate of 0.001. The number of attention head is implemented with 4. ($w_{mse}$, $w_{\mathrm{KL}}$) are set to (100, 0.001). For MDM, we train it for 2000 epochs on BABEL and NTU RGB-D, and \guPr{for} 1000 epochs \guPr{on} HumanAct12, with \guPr{a} learning rate \guPr{of} 0.001. The total (reverse) diffusion is called for $T = 1000$ iterations. For DS, we set $(w_{div}, w_{\mathrm{KL}}^{samp})$ to (200,1) with \guPr{a} learning rate of 0.01 for 30 epochs. The AinB-VAE decoder is kept frozen during the sampler training. All the experiments are conducted on RTX4090.  

\section{Conclusion}
We have proposed a framework, CDFace, to enable the synthesis of stochastic facial motions driven by speech signals, even on small-scale datasets. Motivated by the diversity-promoting loss, CDFace learns to predict a set of facial latent codes whose decoded movements are richly diversified. We also incorporate a masking operation in the lip region such that the diversification can yielded in an audio-faithful manner. To further allow control over facial parts, we individually prepare the facial prior for each of them, and then predict different facial portions sequentially to compose the entire face. CDFace unifies generation diversity and controllability of facial animation into one formulation. Extensive experimental results demonstrate the state-of-the-art effectiveness of facial motion synthesis, regarding diversity, realism, and expressiveness.

% \fi

%\appendices

%\section{}


\ifCLASSOPTIONcaptionsoff
  \newpage
\fi

\bibliographystyle{abbrv}
\bibliography{reference}

\begin{thebibliography}{10}

\bibitem{baevski2020wav2vec}
A.~Baevski, Y.~Zhou, A.~Mohamed, and M.~Auli.
\newblock wav2vec 2.0: A framework for self-supervised learning of speech representations.
\newblock {\em Advances in neural information processing systems}, 33:12449--12460, 2020.

\bibitem{cao2005expressive}
Y.~Cao, W.~C. Tien, P.~Faloutsos, and F.~Pighin.
\newblock Expressive speech-driven facial animation.
\newblock {\em ACM Transactions on Graphics (TOG)}, 24(4):1283--1302, 2005.

\bibitem{chai2022personalized}
Y.~Chai, T.~Shao, Y.~Weng, and K.~Zhou.
\newblock Personalized audio-driven 3d facial animation via style-content disentanglement.
\newblock {\em IEEE Transactions on Visualization and Computer Graphics}, 30(3):1803--1820, 2022.

\bibitem{cudeiro2019capture}
D.~Cudeiro, T.~Bolkart, C.~Laidlaw, A.~Ranjan, and M.~J. Black.
\newblock Capture, learning, and synthesis of 3d speaking styles.
\newblock In {\em Proceedings of the IEEE/CVF conference on computer vision and pattern recognition}, pages 10101--10111, 2019.

\bibitem{edwards2016jali}
P.~Edwards, C.~Landreth, E.~Fiume, and K.~Singh.
\newblock Jali: an animator-centric viseme model for expressive lip synchronization.
\newblock {\em ACM Transactions on graphics (TOG)}, 35(4):1--11, 2016.

\bibitem{esser2021taming}
P.~Esser, R.~Rombach, and B.~Ommer.
\newblock Taming transformers for high-resolution image synthesis.
\newblock In {\em Proceedings of the IEEE/CVF conference on computer vision and pattern recognition}, pages 12873--12883, 2021.

\bibitem{fan2022faceformer}
Y.~Fan, Z.~Lin, J.~Saito, W.~Wang, and T.~Komura.
\newblock Faceformer: Speech-driven 3d facial animation with transformers.
\newblock In {\em Proceedings of the IEEE/CVF Conference on Computer Vision and Pattern Recognition}, pages 18770--18780, 2022.

\bibitem{fanelli20103}
G.~Fanelli, J.~Gall, H.~Romsdorfer, T.~Weise, and L.~Van~Gool.
\newblock A 3-d audio-visual corpus of affective communication.
\newblock {\em IEEE Transactions on Multimedia}, 12(6):591--598, 2010.

\bibitem{huang2022unicolor}
Z.~Huang, N.~Zhao, and J.~Liao.
\newblock Unicolor: A unified framework for multi-modal colorization with transformer.
\newblock {\em ACM Transactions on Graphics (TOG)}, 41(6):1--16, 2022.

\bibitem{karras2017audio}
T.~Karras, T.~Aila, S.~Laine, A.~Herva, and J.~Lehtinen.
\newblock Audio-driven facial animation by joint end-to-end learning of pose and emotion.
\newblock {\em ACM Transactions on Graphics (ToG)}, 36(4):1--12, 2017.

\bibitem{kim2024motion}
H.~J. Kim and E.~Ohn-Bar.
\newblock Motion diversification networks.
\newblock In {\em Proceedings of the IEEE/CVF Conference on Computer Vision and Pattern Recognition}, pages 1650--1660, 2024.

\bibitem{li2024pose}
B.~Li, X.~Wei, B.~Liu, Z.~He, J.~Cao, and Y.-K. Lai.
\newblock Pose-aware 3d talking face synthesis using geometry-guided audio-vertices attention.
\newblock {\em IEEE Transactions on Visualization and Computer Graphics}, 2024.

\bibitem{li2017learning}
T.~Li, T.~Bolkart, M.~J. Black, H.~Li, and J.~Romero.
\newblock Learning a model of facial shape and expression from 4d scans.
\newblock {\em ACM Trans. Graph.}, 36(6):194--1, 2017.

\bibitem{lin2024emoface}
Y.~Lin, L.~Peng, J.~Hu, X.~Li, W.~Kang, S.~Lei, X.~Wu, and H.~Xu.
\newblock Emoface: Emotion-content disentangled speech-driven 3d talking face with mesh attention.
\newblock {\em arXiv preprint arXiv:2408.11518}, 2024.

\bibitem{lin2024glditalker}
Y.~Lin, L.~Xiong, X.~Li, W.~Kang, X.~Wu, L.~Peng, S.~Lei, H.~Xu, and Z.~Fan.
\newblock Glditalker: Speech-driven 3d facial animation with graph latent diffusion transformer.
\newblock {\em arXiv preprint arXiv:2408.01826}, 2024.

\bibitem{liu2024emage}
H.~Liu, Z.~Zhu, G.~Becherini, Y.~Peng, M.~Su, Y.~Zhou, X.~Zhe, N.~Iwamoto, B.~Zheng, and M.~J. Black.
\newblock Emage: Towards unified holistic co-speech gesture generation via expressive masked audio gesture modeling.
\newblock In {\em Proceedings of the IEEE/CVF Conference on Computer Vision and Pattern Recognition}, pages 1144--1154, 2024.

\bibitem{liu2021geometry}
J.~Liu, B.~Hui, K.~Li, Y.~Liu, Y.-K. Lai, Y.~Zhang, Y.~Liu, and J.~Yang.
\newblock Geometry-guided dense perspective network for speech-driven facial animation.
\newblock {\em IEEE Transactions on Visualization and Computer Graphics}, 28(12):4873--4886, 2021.

\bibitem{loshchilov2017decoupled}
I.~Loshchilov and F.~Hutter.
\newblock Decoupled weight decay regularization.
\newblock {\em arXiv preprint arXiv:1711.05101}, 2017.

\bibitem{mao2021generating}
W.~Mao, M.~Liu, and M.~Salzmann.
\newblock Generating smooth pose sequences for diverse human motion prediction.
\newblock In {\em Proceedings of the IEEE/CVF International Conference on Computer Vision}, pages 13309--13318, 2021.

\bibitem{massaro2012animated}
D.~Massaro, M.~Cohen, R.~Clark, M.~Tabain, and J.~Beskow.
\newblock Animated speech: Research progress and applications.
\newblock 2012.

\bibitem{ng2022learning}
E.~Ng, H.~Joo, L.~Hu, H.~Li, T.~Darrell, A.~Kanazawa, and S.~Ginosar.
\newblock Learning to listen: Modeling non-deterministic dyadic facial motion.
\newblock In {\em Proceedings of the IEEE/CVF Conference on Computer Vision and Pattern Recognition}, pages 20395--20405, 2022.

\bibitem{parke2008computer}
F.~I. Parke and K.~Waters.
\newblock {\em Computer facial animation}.
\newblock CRC press, 2008.

\bibitem{pei2006transferring}
Y.~Pei and H.~Zha.
\newblock Transferring of speech movements from video to 3d face space.
\newblock {\em IEEE Transactions on Visualization and Computer Graphics}, 13(1):58--69, 2006.

\bibitem{peng2021generating}
J.~Peng, D.~Liu, S.~Xu, and H.~Li.
\newblock Generating diverse structure for image inpainting with hierarchical vq-vae.
\newblock In {\em Proceedings of the IEEE/CVF conference on computer vision and pattern recognition}, pages 10775--10784, 2021.

\bibitem{razavi2019generating}
A.~Razavi, A.~Van~den Oord, and O.~Vinyals.
\newblock Generating diverse high-fidelity images with vq-vae-2.
\newblock {\em Advances in neural information processing systems}, 32, 2019.

\bibitem{richard2021meshtalk}
A.~Richard, M.~Zollh{\"o}fer, Y.~Wen, F.~De~la Torre, and Y.~Sheikh.
\newblock Meshtalk: 3d face animation from speech using cross-modality disentanglement.
\newblock In {\em Proceedings of the IEEE/CVF International Conference on Computer Vision}, pages 1173--1182, 2021.

\bibitem{stan2023facediffuser}
S.~Stan, K.~I. Haque, and Z.~Yumak.
\newblock Facediffuser: Speech-driven 3d facial animation synthesis using diffusion.
\newblock In {\em Proceedings of the 16th ACM SIGGRAPH Conference on Motion, Interaction and Games}, pages 1--11, 2023.

\bibitem{sun2024diffpose}
Z.~Sun, T.~Lv, S.~Ye, M.~Lin, J.~Sheng, Y.-H. Wen, M.~Yu, and Y.-J. Liu.
\newblock Diffposetalk: Speech-driven stylistic 3d facial animation and head pose generation via diffusion models.
\newblock {\em ACM Trans. Graph.}, 43(4), jul 2024.

\bibitem{sung2024laughtalk}
K.~Sung-Bin, L.~Hyun, D.~H. Hong, S.~Nam, J.~Ju, and T.-H. Oh.
\newblock Laughtalk: Expressive 3d talking head generation with laughter.
\newblock In {\em Proceedings of the IEEE/CVF Winter Conference on Applications of Computer Vision}, pages 6404--6413, 2024.

\bibitem{tan2024flowvqtalker}
S.~Tan, B.~Ji, and Y.~Pan.
\newblock Flowvqtalker: High-quality emotional talking face generation through normalizing flow and quantization.
\newblock In {\em Proceedings of the IEEE/CVF Conference on Computer Vision and Pattern Recognition}, pages 26317--26327, 2024.

\bibitem{taylor2017deep}
S.~Taylor, T.~Kim, Y.~Yue, M.~Mahler, J.~Krahe, A.~G. Rodriguez, J.~Hodgins, and I.~Matthews.
\newblock A deep learning approach for generalized speech animation.
\newblock {\em ACM Transactions On Graphics (TOG)}, 36(4):1--11, 2017.

\bibitem{taylor2012dynamic}
S.~L. Taylor, M.~Mahler, B.-J. Theobald, and I.~Matthews.
\newblock Dynamic units of visual speech.
\newblock In {\em Proceedings of the 11th ACM SIGGRAPH/Eurographics conference on Computer Animation}, pages 275--284, 2012.

\bibitem{thambiraja20233diface}
B.~Thambiraja, S.~Aliakbarian, D.~Cosker, and J.~Thies.
\newblock 3diface: Diffusion-based speech-driven 3d facial animation and editing.
\newblock {\em arXiv preprint arXiv:2312.00870}, 2023.

\bibitem{thambiraja2023imitator}
B.~Thambiraja, I.~Habibie, S.~Aliakbarian, D.~Cosker, C.~Theobalt, and J.~Thies.
\newblock Imitator: Personalized speech-driven 3d facial animation.
\newblock In {\em Proceedings of the IEEE/CVF International Conference on Computer Vision}, pages 20621--20631, 2023.

\bibitem{van2017neural}
A.~Van Den~Oord, O.~Vinyals, et~al.
\newblock Neural discrete representation learning.
\newblock {\em Advances in neural information processing systems}, 30, 2017.

\bibitem{xing2023codetalker}
J.~Xing, M.~Xia, Y.~Zhang, X.~Cun, J.~Wang, and T.-T. Wong.
\newblock Codetalker: Speech-driven 3d facial animation with discrete motion prior.
\newblock In {\em Proceedings of the IEEE/CVF Conference on Computer Vision and Pattern Recognition}, pages 12780--12790, 2023.

\bibitem{xu2022diverse}
S.~Xu, Y.-X. Wang, and L.-Y. Gui.
\newblock Diverse human motion prediction guided by multi-level spatial-temporal anchors.
\newblock In {\em European Conference on Computer Vision}, pages 251--269. Springer, 2022.

\bibitem{xu2013practical}
Y.~Xu, A.~W. Feng, S.~Marsella, and A.~Shapiro.
\newblock A practical and configurable lip sync method for games.
\newblock In {\em Proceedings of Motion on Games}, pages 131--140, 2013.

\bibitem{yang2024probabilistic}
K.~D. Yang, A.~Ranjan, J.-H.~R. Chang, R.~Vemulapalli, and O.~Tuzel.
\newblock Probabilistic speech-driven 3d facial motion synthesis: New benchmarks methods and applications.
\newblock In {\em Proceedings of the IEEE/CVF Conference on Computer Vision and Pattern Recognition}, pages 27294--27303, 2024.

\bibitem{yi2023generating}
H.~Yi, H.~Liang, Y.~Liu, Q.~Cao, Y.~Wen, T.~Bolkart, D.~Tao, and M.~J. Black.
\newblock Generating holistic 3d human motion from speech.
\newblock In {\em Proceedings of the IEEE/CVF Conference on Computer Vision and Pattern Recognition}, pages 469--480, 2023.

\bibitem{yuvector2022}
J.~Yu, X.~Li, J.~Y. Koh, H.~Zhang, R.~Pang, J.~Qin, A.~Ku, Y.~Xu, J.~Baldridge, and Y.~Wu.
\newblock Vector-quantized image modeling with improved vqgan.
\newblock In {\em International Conference on Learning Representations}, 2022.

\bibitem{yuan2020dlow}
Y.~Yuan and K.~Kitani.
\newblock Dlow: Diversifying latent flows for diverse human motion prediction.
\newblock In {\em Computer Vision--ECCV 2020: 16th European Conference, Glasgow, UK, August 23--28, 2020, Proceedings, Part IX 16}, pages 346--364. Springer, 2020.

\bibitem{zhang2024codebook}
B.~Zhang, H.~Wang, C.~Luo, X.~Li, G.~Liang, Y.~Ye, X.~Qi, and Y.~He.
\newblock Codebook transfer with part-of-speech for vector-quantized image modeling.
\newblock In {\em Proceedings of the IEEE/CVF Conference on Computer Vision and Pattern Recognition}, pages 7757--7766, 2024.

\bibitem{zhang2023generating}
J.~Zhang, Y.~Zhang, X.~Cun, Y.~Zhang, H.~Zhao, H.~Lu, X.~Shen, and Y.~Shan.
\newblock Generating human motion from textual descriptions with discrete representations.
\newblock In {\em Proceedings of the IEEE/CVF conference on computer vision and pattern recognition}, pages 14730--14740, 2023.

\bibitem{zhou2022towards}
S.~Zhou, K.~Chan, C.~Li, and C.~C. Loy.
\newblock Towards robust blind face restoration with codebook lookup transformer.
\newblock {\em Advances in Neural Information Processing Systems}, 35:30599--30611, 2022.

\bibitem{zhou2018visemenet}
Y.~Zhou, Z.~Xu, C.~Landreth, E.~Kalogerakis, S.~Maji, and K.~Singh.
\newblock Visemenet: Audio-driven animator-centric speech animation.
\newblock {\em ACM Transactions on Graphics (TOG)}, 37(4):1--10, 2018.

\bibitem{zollhofer2018state}
M.~Zollh{\"o}fer, J.~Thies, P.~Garrido, D.~Bradley, T.~Beeler, P.~P{\'e}rez, M.~Stamminger, M.~Nie{\ss}ner, and C.~Theobalt.
\newblock State of the art on monocular 3d face reconstruction, tracking, and applications.
\newblock In {\em Computer graphics forum}, volume~37, pages 523--550. Wiley Online Library, 2018.

\end{thebibliography}

\end{document}